\documentclass[runningheads]{llncs}
\usepackage[T1]{fontenc}
\usepackage{graphicx}
\usepackage{booktabs}
\usepackage[misc]{ifsym}
\newcommand{\corr}{(\Letter)}
\usepackage{mwe}
\usepackage{multirow}
\usepackage{tabularx} 
\usepackage{lipsum}
\usepackage{enumitem}
\usepackage{subcaption}
\usepackage{tablefootnote}
\usepackage{color}
\usepackage{threeparttable}
\usepackage{amsmath}
\usepackage{amsfonts}
\newcommand{\name}{PSP}

\begin{document}

\title{PSP: Pre-Training and Structure Prompt Tuning for Graph Neural Networks}

\titlerunning{PSP: Pre-Training and Structure Prompt Tuning for Graph Neural Networks}


\author{Qingqing Ge\inst{1} \and
Zeyuan Zhao\inst{1} \and
Yiding Liu\inst{2} \and Anfeng Cheng\inst{2} \and Xiang Li\inst{1} \corr \and Shuaiqiang Wang\inst{2} \and Dawei Yin\inst{2}}

\authorrunning{Q. Ge et al.}

\institute{School of Data Science and Engineering, East China Normal University, China \\
\email{\{qingqingge, zeyuanzhao\}@stu.ecnu.edu.cn} \\
\email{xiangli@dase.ecnu.edu.cn} \\
\and
Baidu Inc., China \\
\email{\{liuyiding.tanh, anfcheng2, shqiang.wang\}@gmail.com} \\
\email{yindaei@acm.org}}

\tocauthor{Qingqing Ge, Zeyuan Zhao, Yiding Liu, Anfeng Cheng, Xiang Li, Shuaiqiang Wang, Dawei Yin}
\toctitle{PSP: Pre-Training and Structure Prompt Tuning for Graph Neural Networks}

\maketitle              

\begin{abstract}
Graph Neural Networks (GNNs) are powerful in learning semantics of graph data.
Recently, a new paradigm ``pre-train \& prompt'' has shown promising results in adapting GNNs to various tasks with less supervised data. 
The success of such paradigm can be attributed to the 
more consistent objectives of pre-training and task-oriented prompt tuning, where the pre-trained knowledge can be effectively transferred to downstream tasks. 
Most existing methods are based on the class prototype vector framework.
However, 
in the few-shot scenarios, given few labeled data, 
class prototype vectors are difficult to be accurately constructed or learned.
Meanwhile,
the structure information of graph is usually exploited during pre-training for learning node representations, while neglected in the prompt tuning stage for learning more accurate prototype vectors.
In addition, they generally ignore the impact of heterophilous neighborhoods on node representation and are not suitable for heterophilous graphs.
To bridge these gaps, 
we propose a novel pre-training and structure prompt tuning framework for GNNs,
namely \name, which consistently exploits structure information in both pre-training and prompt tuning stages. 
In particular, \name~1) employs a dual-view contrastive learning 
to align the latent semantic spaces of node attributes and
graph structure,
and 2) incorporates structure information in prompted graph to construct more accurate prototype vectors and elicit more pre-trained knowledge in prompt tuning.
We conduct extensive experiments
on node classification and graph classification tasks to evaluate the effectiveness of \name.
We show that \name~can lead to superior performance in few-shot scenarios on both homophilous and heterophilous graphs.
The implemented code is available at 
\url{https://github.com/gqq1210/PSP}.

\end{abstract}

\section{Introduction}

Graph Neural Networks (GNNs) have been widely applied in a variety of fields,
such as social network analysis \cite{hamilton2017inductive}, financial risk control \cite{wang2019semi},
and recommender systems \cite{wu2022graph},
where both structural and attribute information are learned via message passing on the graphs \cite{kipf2016semi}.
Recently,
extensive efforts \cite{hou2022graphmae,li2023seegera}
have been made to design graph pre-training methods,
which are further fine-tuned for various downstream tasks.
Nevertheless,
inconsistent objectives of pre-training and fine-tuning often leads to catastrophic forgetting during downstream adaptation \cite{zhu2023sgl},
especially when the downstream supervised data is too scarce to be easily over-fitted.

To bridge this gap, 
many prompt tuning methods for GNNs \cite{sun2022gppt,liu2023graphprompt,zhu2023sgl,sun2023all,fang2022universal,yu2023generalized,chen2023ultra,yu2023multigprompt,tan2023virtual}~have also been proposed
to achieve remarkable performance in few-shot learning tasks on graphs.
In particular, the key insight of these methods is to freeze the pre-trained model (i.e., GNN) and introduce extra task-specific parameters, which learns to exploit the pre-trained knowledge for downstream tasks. For example, 
GPPT \cite{sun2022gppt} and GraphPrompt \cite{liu2023graphprompt}~pre-train a GNN model based on the link prediction task,
then they take the class prototype vectors and the readout function as parameters respectively to reformulate the downstream node/graph classification task into the same format as link prediction.

\begin{figure}[t]
    \centering
    \includegraphics[width=0.65\textwidth]{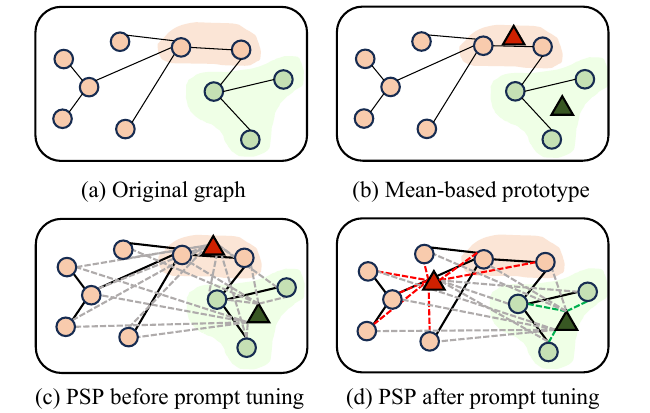}
    \caption{The construction of class prototype vectors. The colored areas contain the labeled nodes for training.
    The circles represent nodes, and the triangles represent class prototype vectors for node classification task.
    The solid black lines and gray dashed lines denote the original edges in the graph and the new weighted edges, respectively.
    For each node, the dashed line in red or green denotes the edges with the largest weight to the class prototype vector.
    }
    \label{fig:intro}
\end{figure}

Despite the initial success,
a clear limitation 
in these existing models
is that graph structure, 
as the key ingredient in pre-training,
is under-explored {when constructing class prototype vectors} in prompt tuning, which limits their effectiveness in unleashing pre-trained knowledge.
In particular, their task-specific parameters (e.g., class prototype vectors or readout functions) are usually learned only with few labeled data. 
They fail to consider the relationships between the task and the massive unlabeled data,
which could also provide rich pre-trained knowledge 
that is very useful for the task at hand. This is even more important when the labeled data is scarce, e.g., few-shot node classification.
As shown in Figure \ref{fig:intro} (b),
existing methods that directly use the average embeddings of labeled nodes/graphs as the class prototype representations can easily be undermined by noisy/outlier data
when the number of labeled nodes is scarce. 
In contrast, facilitating class representation learning with structural connections between class prototype vectors and unlabeled nodes could help solve this issue (as shown in Figure \ref{fig:intro} (d)).
{Moreover, most existing methods are designed for homophilous graphs, 
relying excessively on the graph structural information while disregarding the impact of heterophilous neighborhoods on node representations. 
This adversely affects the model performance 
on heterophilous graphs.}

In this paper,
we propose a novel \textbf{P}re-training and \textbf{S}tructure \textbf{P}rompt tuning (\name) framework,
which unifies the objectives of pre-training and prompt tuning for GNNs
and integrates structural information in both pre-training and prompt tuning stages {to construct more accurate prototype vectors}.
For pre-training,
{inspired by \cite{lim2021large}, 
we separate attribute and structural information,}
employing dual-view contrastive learning to align the latent semantic spaces of node attributes and graph structure.
Specifically,
one view is implemented with MLP,
which only uses 
node attributes in the graph.
The other view adopts GNN
to leverage both node attributes and structural information of the graph.
For downstream prompt-tuning,
we fix the learned parameters of MLP and GNN in the pre-training stage,
add class prototype vectors as new nodes to the raw graph
and introduce structural connections
between prototype vectors and original nodes as prompts
to learn more accurate prototype vectors (see Figure~\ref{fig:intro} (c)).
Note that weights associated with these connections are parameters to be learned.
In the training phase,
we use representations of labeled nodes/graphs calculated by MLP as anchors,
and representations of prototype vectors obtained through GNN as positive/negative samples.
Specifically, the prototype vector in the same class as the anchor is considered as positive sample,
while prototype vectors in other classes serve as negative samples.
Then, contrastive learning between nodes/graphs (MLP-based view) and prototype vectors (GNN-based view) is performed to learn prompt parameters.
As a result,
we unify the objectives of pre-training and prompt tuning. 
After prompt tuning, the nodes and their corresponding class prototypes
are learned to have higher weights on the edges, as shown in Figure~\ref{fig:intro}(d).
We also experimentally show the results in Figure \ref{fig:prompt} of Section \ref{sec:model_analysis}.
Based on the learned weights,
for each prototype vector,
GNN formulates its embedding by weighted-aggregating information from its neighboring nodes, i.e., all the nodes in the raw graph.
This helps learn better prototype vectors by leveraging both labeled nodes and massive unlabeled nodes in the graph, which is particularly useful in few-shot scenarios.
Finally,
in the testing stage,
node/graph classification can be conducted via comparing the similarity of representations between node/graph and the prototype vectors.
Compared with existing graph prompt tuning methods, our method is more desirable in learning better prototype vectors, as we leverage both labeled nodes and massive unlabeled nodes in the graph, which is particularly useful in few-shot scenarios.
We further highlight that 
our prompt tuning method is applicable to both homophilous and heterophilous graphs.
First,
node/graph representations computed from MLP-based view are not affected by structural heterophily.
Second,
prototype vectors calculated from GNN-based view are 
based on the learned weights in structure prompt tuning, 
which takes all nodes in the raw graph as neighbors
and learns to assign large (small) weights to those in the same (different) class.
As such,
the computation of prototype vectors is less affected by graph heterophily.
To summarize, 
our main contributions in this paper are:
\begin{itemize}
    \item
    We propose an effective graph pre-training and prompt tuning framework \name,
    which unifies the objectives of pre-training and prompt tuning.
    \item
    We present a novel prompt tuning strategy,
    which introduces
    a learnable structure prompt to learn high-quality class prototype vectors and 
    enhance model performance on both homophilous and heterophilous graphs.
    \item
    We extensively demonstrate the effectiveness of \name~with different benchmark datasets on both node classification and graph classification.
    In particular, we vary the number of labeled training data and show that \name~can lead to better performance in challenging few-shot scenarios.
\end{itemize}

\section{Related work}



\subsection{Graph Pre-training}
Inspired by the remarkable achievements of pre-trained models in Natural Language Processing (NLP) \cite{long2022vision}~and Computer Vision (CV) \cite{qiu2020pre},
graph pre-training \cite{xia2022survey}~emerges as a powerful paradigm that leverages self-supervision on label-free graphs to learn intrinsic graph properties.
Some effective and commonly-used pre-training strategies include node-level comparison~\cite{zhu2021graph}, edge-level pretext~\cite{jin2020self}, and graph-level contrastive learning~\cite{you2020graph}.
Recently, there are also some newly proposed pre-training methods \cite{hu2020gpt,lu2021learning}.
However, 
these approaches do not consider the gap between pre-training and downstream objectives,
which limits their generalization ability to handle different tasks.

\subsection{Prompt-based Learning}

The training strategy ``pre-train \& fine-tune'' is widely used to adapt pre-trained models onto specific downstream tasks.
However,
this strategy ignores the inherent gap between the objectives of pre-training and diverse downstream tasks, where the knowledge learned via pre-training could be forgotten or ineffectively leveraged for downstream tasks,
leading to poor performance.

To bridge this gap,
NLP proposes a new paradigm,
namely “pre-train \& prompt”.
These methods freeze the parameters of the pre-trained models and introduce additional learnable components in the input space,
thereby enhancing the compatibility between inputs and pre-trained models.
On graph data, 
there are a handful of studies that adopt prompt tuning to learn more generalizable GNNs.
GPPT \cite{sun2022gppt}~relies on edge prediction as the pre-training task and reformulates the downstream task as edge prediction by introducing task tokens
for node classification.
GraphPrompt \cite{liu2023graphprompt}~proposes a unified framework based on subgraph similarity and link prediction,
hinging on a learnable prompt to actively guide downstream tasks using task-specific aggregation in readout function.
and computes class prototype vectors via supervised prototypical contrastive learning.
GPF \cite{fang2022universal} extends the node embeddings with additional task-specific prompt parameters,
and can be applied to the pre-trained GNN models that employ any pre-training strategy.
ProG \cite{sun2023all} reformulates node-level and edge-level tasks to graph-level tasks,
and introduces the meta-learning technique to the graph prompt tuning study.

Despite their success, we observe that
most of them utilize the 
structure information
in pre-training,
while ignoring it in downstream prompt tuning stage for learning more accurate prototype vectors.
This restricts their effectiveness to fully utilize pre-trained knowledge stored in the entire graph.
In particular,
their task-specific parameters are usually learned only with labeled nodes while massive unlabeled nodes are disregarded,
leading to 
poor performance in more challenging few-shot scenarios.
{In addition, they ignore the impact of heterophilous neighborhoods on node representation and are not suitable for heterophilous graphs.}
In this paper,
our proposed \name~employs a dual-view contrastive learning and integrates
structure information in both pre-training and prompt tuning stage {to construct more accurate prototype vectors},
achieving superior performance in few-shot learning tasks on both homophilous
and heterophilous graphs.

\section{Preliminary}
\textbf{Graph.}
We denote a graph as $\mathcal{G}=(\mathcal{V}, \mathcal{E})$,
where $\mathcal{V}=\left \{ v_i \right \} _{i=1}^N$ is a set of $N$ nodes
and $\mathcal{E}\subseteq \mathcal{V} \times \mathcal{V}$ is a set of edges.
We also define $\mathbf{A}\in \mathbb{R}^{N\times N}$ as the adjacency matrix of $\mathcal{G}$, where $A_{ij} = 1$ if $v_i$ and $v_j$ are connected in $\mathcal{G}$, and $A_{ij} = 0$ otherwise.
Each node $v_i$ in the graph is associated with an $F$-dimensional feature vector $\mathbf{x}_i \in \mathbb{R}^{1\times F}$, and the feature matrix of all nodes is defined as $\mathbf{X}\in \mathbb{R}^{N\times F}$.

\noindent \textbf{Research problem.}
In this paper,
we investigate the problem of graph pre-training and prompt tuning,
which learns representations of graph data via pre-training, and transfer the pre-trained knowledge to solve downstream tasks, such as node classification and graph classification.
Moreover, we further consider the scenarios where downstream tasks are given limited supervision, i.e., $k$-shot classification. For each class, only $k$ labeled samples
(i.e., nodes or graphs) are provided as training data.

\noindent \textbf{Prompt tuning of pre-trained models}.
Given a pre-trained model, 
a set of learnable prompt parameters $\theta$ and a labeled task dataset $\mathcal{D}$, 
we fix the parameters of the pre-trained model 
and only optimize $\theta$ with $\mathcal{D}$ for the downstream graph tasks.

\section{Proposed Method}

\begin{figure}[t]
    \centering
    \includegraphics[width=0.86\textwidth]{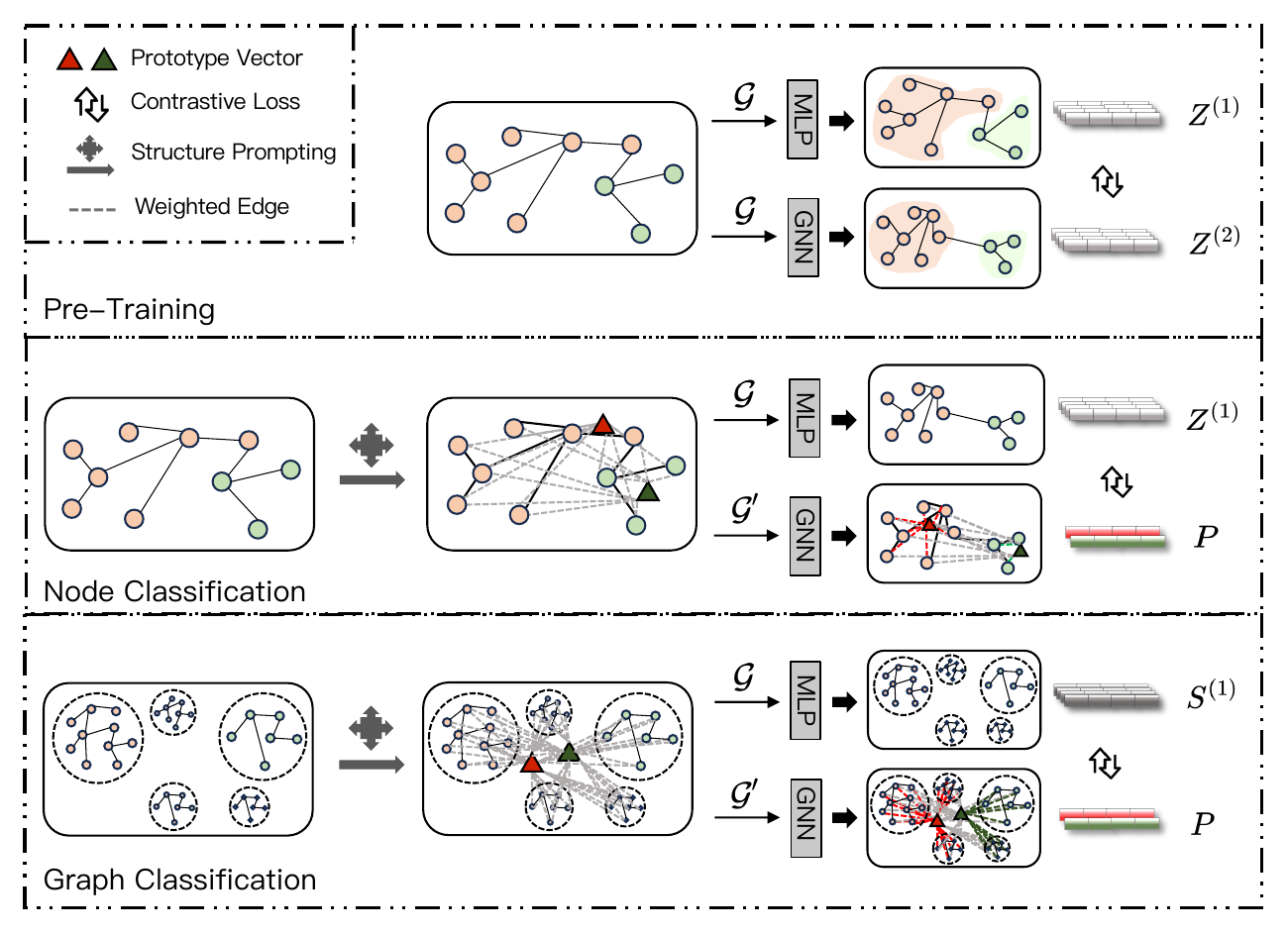}
    \caption{Overall framework of \name.
    Top: pre-training.
    Middle: prompt tuning for node classification.
    Bottom: prompt tuning for graph classification.
    }
    \label{fig:model}
\end{figure}

In this paper, we propose a novel 
graph pre-training and structure prompt tuning framework,
namely \name,
which unifies the objectives of pre-training and prompt tuning for GNNs and integrates structure information in both pre-training and prompt tuning stage to achieve better performance in more challenging  few-shot scenarios.
The overall framework of \name~ is shown in Figure \ref{fig:model}.

\subsection{Graph Pre-training}
For graph data, both node attributes and structural information are critical for revealing the underlying semantics of graph during the pre-training phase.
Inspired by the success of LINKX
\cite{lim2021large} that separately learns node embeddings from attributes and graph structure for heterophilous graphs,
we design a dual-view contrastive learning method to align the latent semantic spaces of node attributes and graph structure.
The upper part of Figure \ref{fig:model} shows the dual-view pre-training paradigm. In particular, one view is implemented with MLP,
which only uses node attributes in the graph, while the other adopts GNN to leverage both node attributes and structural information of the graph. Formally, we define the node representations computed by the two views as:
\begin{equation}
    \mathbf{Z}^{(1)} = \textrm{MLP}(\mathbf{X}) \;\;\;\; \textrm{and} \;\;\;\; \mathbf{Z}^{(2)} = \textrm{GNN}(\mathbf{X}, \mathbf{A}),
\end{equation}
where $\mathbf{Z}^{(1)}\in \mathbb{R}^{N\times D}$ and $\mathbf{Z}^{(2)}\in \mathbb{R}^{N\times D}$ have the same latent dimensionality $D$.

To optimize the MLP and GNN, we leverage a contrastive loss function $\mathcal{L}_{pre}$ to maximize the similarity between the two representations of the same node, denoted as $\mathbf{z}^{(1)}_i$ and $\mathbf{z}^{(2)}_i$ for node $v_i$.
For an anchor $\mathbf{z}^{(1)}_i$,
other node representations $\mathbf{z}^{(2)}_j$ are considered as negative samples.
More specifically, we formulate the loss as the normalized temperature-scaled cross entropy loss \cite{chen2020simple} as
\begin{equation}
\label{eq:pre_loss}
    \mathcal{L}_{pre}=-\frac{1}{N}\sum_{i}^{N}\log\frac{\exp(\mathrm{sim}(\mathbf{z}^{(1)}_i, \mathbf{z}^{(2)}_i)/\tau )}{{\textstyle \sum_{j=0, j\ne i}^{N}} (\exp(\mathrm{sim}(\mathbf{z}^{(1)}_i, \mathbf{z}^{(2)}_j)/\tau))},
\end{equation}
where $N$ is the number of nodes, $\tau$ is a temperature parameter, and $\textrm{sim}(\cdot)$ is implemented with cosine similarity.
In general, the dual-view contrastive pre-training can exploit both attribute and structure information of graph to encode generalizable knowledge in the output node embeddings.
In the next subsection, 
we introduce graph structure prompt tuning that leverages such knowledge in downstream classification tasks.

\subsection{Graph Structure Prompt Tuning}
Next, we demonstrate how we freeze the pre-trained model and adapt it to different downstream tasks on graph. 
In particular, we propose a novel method, namely structure prompt tuning, which considers the structure relationships between the graph data and the task at hand. 
Compared to existing graph prompt tuning methods, 
{the structure relationships in our method allow the task to more effectively leverage the pre-trained knowledge embedded in the graph data}. 
In the following, we elaborate the structure prompt tuning method on two representative tasks, i.e., node classification and graph classification.

\subsubsection{Node Classification}
Our method is based on the prototype-based framework~\cite{liu2023graphprompt,zhu2023sgl}, which learns a prototype embedding vector $\mathbf{p}_c$ for each node class $c \in \{1, 2, ..., C\}$. In particular, our method comprises three steps: 1) structure prompting, 2) prompt initialization, and 3) prompt tuning. 

\noindent\textbf{Step 1: Structure prompting}.
For all the class prototypes, 
the main idea of our method is to consider them as virtual nodes, and connect them to all the original nodes $\mathcal{V}$ in the graph $\mathcal{G}$, as shown in Figure \ref{fig:intro}(c). More specifically, we add a total number of $C$ prototype nodes (denoted as $\mathcal{P}=\left \{ p_c \right \} _{c=1}^C$) and $N\times C$ weighted edges (denoted as $\mathcal{W}$) to construct a prompted graph $\mathcal{G}^\prime$ as
\begin{equation}
    \mathcal{G}^\prime = (\mathcal{V}\cup \mathcal{P}, \mathcal{E}\cup \mathcal{W}).
\end{equation}
\noindent\textbf{Step 2: Prompt initialization}.
Next, we define the attributes and edge weights for the prototype nodes. 
In particular, we simply initialize the attributes for each prototype $p_c$ as the averaged attribute vector of labeled nodes in class $c$: 
\begin{equation}
    \mathbf{x}_c =\frac{1}{|\mathcal{D}_c|}\sum_{(v_i, c)\in \mathcal{D}_c}\mathbf{x}_{i},
\end{equation}
where $\mathcal{D}_c \subseteq \mathcal{D}$ denotes labeled nodes of class $c$ in the training set.
For the newly added edges that connect $\mathcal{P}$ and $\mathcal{V}$, we adopt dot product to initialize their weights $\mathbf{W}\in\mathbb{R}^{N\times C}$ as:
\begin{equation}
    \label{eq:w_init}
    \mathbf{W} = \mathbf{Z}^{(2)}\mathbf{P}^{\mathsf{T}}.
\end{equation}
where $\mathbf{Z}^{(2)}$ denotes node embeddings derived from the pre-trained GNN and
$\mathbf{P}$ represents the prototype vectors with the $c$-th row
$\mathbf{p}_c =\frac{1}{|\mathcal{D}_c|}\sum_{(v_i, c)\in \mathcal{D}_c}\mathbf{z}^{(2)}_{i}$.
During the subsequent prompt tuning step, the parameters of the pre-trained model are frozen, and the weight matrix $\mathbf{W}$ is considered as the only task-specific parameter to be learned.

\noindent\textbf{Step 3: Prompt tuning}.
We conduct prompt tuning on the prompted graph $\mathcal{G}^\prime$ using the same form of contrastive loss function as in Equation~\ref{eq:pre_loss}, while keeping the node representations fixed and only optimize the prototype embeddings. 
Formally, given a labeled dataset $\mathcal{D}$ for the downstream task,
the prompt tuning loss is defined as follows: 
\begin{equation}
    \mathcal{L}_{pro}=-\frac{1}{|\mathcal{D}|}\sum_{(v_i,c_i)\in \mathcal{D}} \log\frac{\exp(\mathrm{sim}(\mathbf{z}^{(1)}_i, \mathbf{p}_{c_i})/\tau )}{\textstyle \sum_{c_j\in C, c_j\ne c_i} \exp(\mathrm{sim}(\mathbf{z}^{(1)}_i, \mathbf{p}_{c_j})/\tau)},
    \label{equ:loss}
\end{equation}
where $\mathbf{p}_c$ is parameterized by $\mathbf{W}$ and we have:
\begin{equation}
    \mathbf{P} = \textrm{GNN}([\mathbf{X}, \mathbf{X}_P], [\mathbf{A}, \mathbf{W}]).
\end{equation}
Here, $\mathbf{X}$ represents the original node features in graph $\mathcal{G}$, $\mathbf{X}_P\in \mathbb{R}^{C\times F}$ represents the prototype features, 
and $[\mathbf{A}, \mathbf{W}] \in \mathbb{R}^{N \times (N+C)}$ represents the new adjacency matrix of $\mathcal{G}^\prime$.
Notably, unlike conventional methods that {directly consider $\mathbf{P}$ as learnable parameters}, we parameterize $\mathbf{P}$ with their structural connections with the graph data, i.e., the added edges $\mathbf{W}$. In other words, only $\mathbf{W}$ would be optimized as parameters when minimizing $\mathcal{L}_{pro}$, which learns to aggregate pre-trained knowledge from all the nodes in the graph to formulate prototype embeddings.

\subsubsection{Graph Classification}
Our method can also be adapted to graph classification with minor changes in each step. In \textbf{structure prompting}, for each graph instance $\mathcal{G}_i$ and a prototype node $p_c$, the added edges between any $v_i \in \mathcal{V}_i$ and $p_c$ share the same weight. As such, the prompt tuning of graph classification also has $N \times C$ parameters,
where $N$ is the number of graphs here.
In \textbf{prompt initialization}, 
we introduce an average-based readout function on node level representations $\mathbf{Z}^{(1)}$ and $\mathbf{Z}^{(2)}$ to compute the graph-level representation $\mathbf{S}^{(1)}$ and $\mathbf{S}^{(2)}$, respectively, as shown in Figure \ref{fig:model}. Then, we can use Equation \ref{eq:w_init} to initialize $\mathbf{W}$ with $\mathbf{Z}^{(2)}$ replaced by $\mathbf{S}^{(2)}$.
In \textbf{prompt tuning}, we also replace $\mathbf{Z}^{(1)}$ by $\mathbf{S}^{(1)}$ in Equation \ref{equ:loss} for optimizing graph-level classification tasks.

\subsubsection{Remarks}
In our proposed \name~method, it worth noting that the weights of pre-trained model are frozen for downstream tasks,
and the prompt tuning is parameterized by the learnable adjacency matrix $\mathbf{W}\in \mathbb{R}^{N\times C}$. Compared with most existing studies where the prototype vectors are directly optimized on few labeled data, this allows the prototype vectors to aggregate pre-trained knowledge from massive unlabeled nodes for more effective task adaptation.

\subsection{Inference}
In the inference stage, classification is performed by comparing the similarity of representations between node/graph (from MLP-based view) and prototype vectors (from GNN-based view) from different classes.
The class corresponding to the prototype vector with the largest similarity is taken as the final prediction of the node/graph.
We use node classification as an example to explain in detail.
By comparing the node representation with each class prototype vector $\mathbf{p}_c$,
we can get the predicted class probability by:
\begin{equation}
    p(c|v_i)= \frac{\exp(\mathrm{sim}(\mathbf{z}^{(1)}_i, \mathbf{p}_{c})/\tau )}{\textstyle \sum_{c=0}^C \exp(\mathrm{sim}(\mathbf{z}^{(1)}_i, \mathbf{p}_c)/\tau)},
    \label{eq:infer}
\end{equation}
where the highest-scored class is chosen as the prediction.
From Equation \ref{eq:infer},
we see that the computation aligns well with objectives in Equations~\ref{eq:pre_loss} and \ref{equ:loss}.


\section{Experiments}
In this section, we conduct extensive experiments on node classification and graph classification tasks with 11 benchmark datasets to evaluate \name.

\subsection{Experimental Setup}
\textbf{Datasets.}
We evaluate the performance of \name\ using various benchmark datasets with diverse properties, including  homophilous graphs \cite{kipf2016semi,hu2020open}: Cora, CiteSeer, PubMed, ogbn-arxiv,
and heterophilous graphs \cite{pei2020geom,liu2023graphprompt,morris2020tudataset}: Chameleon, Actor, ENZYMES, PROTEINS, COX2, BZR, COLLAB.
We summarize these datasets in Table \ref{dataset}.
Note that the “Task” row indicates the type of downstream task, where “N” represents node classification and “G” represents graph classification.

\begin{table}[t]
  \caption{Statistics of the datasets}
  \label{dataset}
  \centering
  \resizebox{\linewidth}{!}{
  \begin{tabular}{c|ccccccccccc}
    \toprule
     & Cora & CiteSeer & PubMed & ogbn-arxiv & Chameleon & Actor & ENZYMES & PROTEINS & COX2 & BZR & COLLAB\\
    \midrule
    Graph & 1 & 1 & 1 & 1 & 1 & 1 & 600 & 1,113 & 467 & 405 & 5000\\
    Graph classes & - & - & - & - & - & - & 6 & 2 & 2 & 2 & 3\\
    Avg. nodes & 2,708 & 3,327 & 19,717 & 169,343 & 2,277 & 7,600 & 32.63 & 39.06 & 41.22 & 35.75 & 74.49\\
    Avg. edges & 5,429 & 4,732 & 44,338 & 1,166,243 & 31,421 & 26,752 & 62.14 & 72.82 & 43.45 & 38.36 & 2457.78\\
    Node features & 1433 & 3703 & 500 & 128 & 2325 & 931 & 18 & 1 & 3 & 3 & 367\\
    Node classes & 7 & 6 & 3 & 40 & 5 & 5 & 3 & 3 & - & - & -\\
    Task (N/G) & N & N & N & N & N & N & N,G & G & G & G & G\\
    \bottomrule
  \end{tabular}}
\end{table}

\noindent \textbf{Baselines.}
To evaluate the proposed \name, we compare it with {4} categories of state-of-the-art approaches as follows.
\textbf{Supervised models}: GCN \cite{kipf2016semi}~and GAT \cite{velivckovic2017graph}.
They use the labeled data to learn GNNs, which are then directly applied for the classification tasks.
\textbf{Graph pre-training models}: EdgeMask \cite{velivckovic2018deep} and GraphCL \cite{you2020graph}.
Following the transfer learning strategy of “pre-train \& fine-tune”, the pre-trained models are fine-tuned on the downstream tasks.
\textbf{Graph few-shot learning models}:
CGPN \cite{wan2021contrastive}~and Meta-PN \cite{ding2022meta}. They are specially designed for few-shot scenarios.
\textbf{Graph prompt models}: GPPT \cite{sun2022gppt}, GraphPrompt \cite{liu2023graphprompt}, GPF \cite{fang2022universal}~and ProG \cite{sun2023all}.
They adopt the “pre-train \& prompt” paradigm, where the pre-trained models are frozen, and task-specific learnable prompts are introduced and trained in the downstream tasks.
Note that we use GraphCL as the pre-training model for GPF.
We also notice that SGL-PT \cite{zhu2023sgl}~and VNT \cite{tan2023virtual}~are recent methods in the similar topic. However, both of them do not release their codes. For fairness, we do not take them as baselines.

\noindent \textbf{Implementation details.}
To train \name, we adopt the Adam optimizer \cite{kingma2014adam}, where the learning rate and weight decay in the pre-training stage are fixed as 1e-4. We set the number of both graph neural layers and multilayer perceptron layers as 2.
We set the hidden dimension for node classification as 128, and for graph classification as 32.
Other hyper-parameters are fine-tuned on the validation set by grid search.
In the prompt tuning stage,
the learning rate is adjusted within \{0.0001, 0.001, 0.01, 0.1\},
the weight decay is chosen from \{1e-5, 1e-4, 1e-3, 1e-2\} and the dropout rate is selected from the range [0.2, 0.8].
Further,
for those non-prompt-based competitors,
some of their results are directly reported from \cite{sun2022gppt}~and \cite{liu2023graphprompt}~(i.e., node classification with 50\% masking ratio and graph classification).
For other cases and prompt-based models,
we fine-tune hyper-parameters with the codes released by their original authors.
For fair comparison,
we report the average results with standard deviations of 5 runs for node classification experiments, 
while the setting of graph classification experiments follows \cite{liu2023graphprompt}.
We run all the experiments on a server with 32G memory and a single Tesla V100 GPU.

\begin{table}[t]
  \caption{Accuracy (\%) on node classification with masking ratio of 50\%.
  We highlight the best score on each dataset in bold. OOM denotes the out-of-the-memory error.}
  \label{tab:node-50}
  \centering
  \resizebox{\linewidth}{!}{
  \begin{tabular}{c|cccc|cc}
    \toprule
    Methods & Cora & CiteSeer & PubMed & ogbn-arxiv & Chameleon & Actor \\
    \midrule
    GCN & 71.78 $\pm$ 0.50 & 52.15 $\pm$ 0.27 & 65.05 $\pm$ 0.15 & 64.19 $\pm$ 0.59 & 30.38 $\pm$ 1.21 & 18.67 $\pm$ 1.94 \\
    GAT & 74.94 $\pm$ 1.26 & 59.50 $\pm$ 0.61 & 69.30 $\pm$ 0.97 & 63.97 $\pm$ 0.69 & 29.14 $\pm$ 0.79 & 20.85 $\pm$ 1.37 \\
    \midrule
    EdgeMask & 76.38 $\pm$ 0.89 & 65.49 $\pm$ 0.90 & 71.29 $\pm$ 0.66 & 64.86 $\pm$ 0.67 & 30.89 $\pm$ 1.15 & 21.76 $\pm$ 0.95 \\
    GraphCL & 76.73 $\pm$ 0.91 & 65.94 $\pm$ 1.20 & 72.03 $\pm$ 1.54 & 65.87 $\pm$ 0.82 & 27.37 $\pm$ 1.40 & 22.18 $\pm$ 1.24 \\
    \midrule
    CGPN & 74.62 $\pm$ 0.99 & 65.32 $\pm$ 1.17 & 67.38 $\pm$ 1.43 & OOM & 31.12 $\pm$ 2.07 & 21.91 $\pm$ 0.39\\
    Meta-PN & 76.89 $\pm$ 1.05 & 65.80 $\pm$ 1.13 & 69.75 $\pm$ 1.11 & 57.37 $\pm$ 0.79 & 30.08 $\pm$ 1.25 & 19.16 $\pm$ 1.04\\
    \midrule
    GPPT & 77.16 $\pm$ 1.35 & 65.81 $\pm$ 0.97 & 72.23 $\pm$ 1.22 & 66.13 $\pm$ 0.44 & 31.28 $\pm$ 0.65 & 22.07 $\pm$ 0.81 \\
    GraphPrompt & 68.43 $\pm$ 0.58 & 61.11 $\pm$ 1.24 & 72.63 $\pm$ 1.72 & 59.13 $\pm$ 0.59 & 31.67 $\pm$ 1.19 & 21.11 $\pm$ 1.35 \\
    GPF & 62.20 $\pm$ 0.93 & 60.78 $\pm$ 1.17 & 68.81 $\pm$ 0.95 & 56.04 $\pm$ 0.97 & 29.31 $\pm$ 1.68 & 20.56 $\pm$ 1.09 \\
    ProG & 72.49 $\pm$ 1.04 & 65.33 $\pm$ 0.93 & 73.70 $\pm$ 1.49 & 67.80 $\pm$ 1.65 & 31.75 $\pm$ 1.25 & 22.82 $\pm$ 1.03 \\
    \midrule
    \name~ & \textbf{77.62 $\pm$ 1.29} & \textbf{67.52 $\pm$ 0.95} & \textbf{75.94 $\pm$ 1.06} & \textbf{69.20 $\pm$ 0.73} & \textbf{34.08 $\pm$ 1.26} & \textbf{25.47 $\pm$ 0.82}\\
    \bottomrule
  \end{tabular}}
\end{table}

\subsection{Node classification}
\label{sec:few-shot}
\noindent\textbf{Experimental setting}.
For homophilous graph datasets,
we use the official splitting of training/validation/testing \cite{kipf2016semi}.
For heterophilous graph datasets,
we randomly sample 20 nodes per class as training set and validation set, respectively.
The remaining nodes which are not sampled will be used for evaluation.
Following the setting of \cite{sun2022gppt},
we randomly mask 50\% of the training labels,
which corresponds to 10-shot for datasets except ogbn-arxiv.

\noindent \textbf{Results.} Table \ref{tab:node-50} summarizes the results, from which we see that:

(1) Supervised learning methods generally perform worse than pre-training methods and prompt methods.
This is because the annotations required by supervised frameworks are not enough.
In contrast,
pre-training approaches are usually facilitated with more prior knowledge,
alleviating the need for labeled data.
However,
these pre-training methods still face an inherent gap between the training objectives of pre-training and downstream tasks.
Pre-trained models may suffer from catastrophic forgetting during downstream adaptation.
Therefore, we can find that compared with pre-training approaches,
prompt-based methods usually achieve better performance.

(2) Our proposed \name~outperforms all the baselines on node classification.
This is because \name~bridges the gap between the pre-training stage and the down-stream prompt tuning stage, leveraging graph structure prompt to provide more information from the massive unlabeled nodes and being applicable to both homophilous and heterophilous graphs. 

\begin{table}[t]
  \caption{Accuracy (\%) on few-shot node classification.}
  \label{tab:node-3}
  \centering
  \resizebox{\linewidth}{!}{
  \begin{tabular}{c|cccc|cc}
  \toprule
  Methods & Cora & CiteSeer & PubMed & ENZYMES & Chameleon & Actor \\
  \midrule
    GCN & 57.83 $\pm$ 5.90 & 49.69 $\pm$ 4.39 & 63.16 $\pm$ 4.56 & 61.49 $\pm$ 12.87 & 27.88  $\pm$ 5.77 & 20.69 $\pm$ 2.96\\
    GAT & 60.34 $\pm$ 4.15 & 52.85 $\pm$ 3.69 & 64.26 $\pm$ 3.17 & 59.94 $\pm$ 2.86 & 26.97 $\pm$ 4.85 & 20.93 $\pm$ 2.67\\
    \midrule
    EdgeMask & 64.10 $\pm$ 2.79 & 55.23 $\pm$ 3.17 & 65.89 $\pm$ 4.26 & 56.17 $\pm$ 14.39 & 23.76 $\pm$ 3.74 & 18.03 $\pm$ 2.48 \\
    GraphCL & 65.89 $\pm$ 3.45 & 58.37 $\pm$ 4.74 & 69.06 $\pm$ 3.24 & 58.73 $\pm$ 16.47 & 22.25 $\pm$ 3.14 & 19.56 $\pm$ 1.15\\
    \midrule
    CGPN & 66.73 $\pm$ 2.87 & 57.14 $\pm$ 3.75 & 65.68 $\pm$ 2.38 & 66.52 $\pm$ 16.18 & 27.17 $\pm$ 2.16 & 20.68 $\pm$ 0.99\\
    Meta-PN & 66.65 $\pm$ 3.15 & 58.20 $\pm$ 2.94 & 67.18 $\pm$ 3.12 & 55.92 $\pm$ 13.28 & 25.83 $\pm$ 2.64 & 18.31 $\pm$ 1.95\\
    \midrule
    GPPT & 64.55 $\pm$ 3.72 & 55.63 $\pm$ 2.55 & 70.07 $\pm$ 6.07 & 53.79 $\pm$ 17.46 & 28.91 $\pm$ 3.23 & 20.88 $\pm$ 1.69 \\
    GraphPrompt & 63.91 $\pm$ 2.43 & 53.42 $\pm$ 4.98 & 68.93 $\pm$ 3.93 & 67.04 $\pm$ 11.48 & 26.35 $\pm$ 3.50 & 20.50 $\pm$ 2.45 \\
    GPF & 63.52 $\pm$ 5.39 & 54.31 $\pm$ 5.21 & 63.98 $\pm$ 3.54 & 60.13 $\pm$ 15.37 & 27.38 $\pm$ 3.62 & 19.32 $\pm$ 2.52 \\
    ProG & 65.68 $\pm$ 4.29 & 59.07 $\pm$ 2.73 & 64.57 $\pm$ 3.81 & 57.22 $\pm$ 17.41 & 29.18 $\pm$ 4.53 & 21.43 $\pm$ 3.27 \\
    \midrule
    \name~ & \textbf{68.65 $\pm$ 2.17} & \textbf{61.7 $\pm$ 4.21} & \textbf{72.23 $\pm$ 4.20} & \textbf{72.86 $\pm$ 14.58} & \textbf{33.23 $\pm$ 3.80} & \textbf{24.74 $\pm$ 2.79}\\
    \bottomrule
  \end{tabular}}
\end{table}

\subsection{Few-shot node classification}
\noindent\textbf{Experimental setting}. To explore more challenging few-shot node classification settings,
we assign a much smaller number of labeled data as the training data for each class.
Specifically,
for ENZYMES,
we follow existing study \cite{liu2023graphprompt}~to only choose graphs that consist of more than 50 nodes,
{which ensures} there are sufficient labeled nodes for testing.
On each graph,
we randomly sample 1 node per class for training and validation, respectively.
The remaining nodes which are not sampled will be used for testing.
For Cora, CiteSeer and PubMed,
we randomly sample 3 nodes per class for training,
while the validation and test sets follow the official splitting \cite{kipf2016semi}.
For Chameleon and Actor,
we randomly sample 3 nodes per class for training and validation respectively,
while the remaining nodes which are not sampled are used for testing.

\noindent \textbf{Results.} Table \ref{tab:node-3} illustrates the results. From the table, we see that:

\noindent (1) In the extremely few-shot scenarios, \name~can still achieve superior performance.
Taking ENZYMES as an example,
the accuracy of \name~is 5.82\% higher than the runner-up.
This is because \name~uses graph structure information to optimize the prototype vectors,
so when the training set is extremely small,
the prototype vectors can also be accurate by capturing the underlying information of other unlabeled nodes.
In contrast,
{few-shot learning models may suffer from inaccurate pseudo labels and}
other methods fail to consider the relationship between the task and the massive unlabeled data.

\noindent (2) Our proposed \name~achieves larger improvements in heterophilous graphs.
\name~achieves a largest improvement of 2.97\% in homogeneous graphs while 5.82\% in heterophilous graphs.
This is because in the downstream prompt tuning stage of \name,
node representations computed from MLP-based view are not affected by the structural
heterophily.
Also, the graph structure prompt reduces the adverse influence from graph heterophily.
Therefore, the prototype vector calculated from GNN-based view are more accurate.
In contrast,
other methods are susceptible to structure heterophily and can only achieve sub-optimal results.

\subsection{Few-shot graph classification}
Following the setting of \cite{liu2023graphprompt},
we conduct 5-shot tasks.
The results are listed in Table \ref{tab:graph class},
from which we observe that
our proposed \name~significantly outperforms the baselines on these datasets.
This again demonstrates the effectiveness of our proposed method.
Notably,
as both node and graph classification tasks share the same pre-trained model on ENZYMES,
the superior performance of \name~on both types of tasks further demonstrates that
the gap between different tasks is better addressed by our unified framework. 

\begin{table}[t]
  \caption{Accuracy (\%) on graph classification.}
  \label{tab:graph class}
  \centering
  \resizebox{\linewidth}{!}{
  \begin{threeparttable}
  \begin{tabular}{c|ccccc}
    \toprule
    Methods & ENZYMES & PROTEINS & COX2 & BZR & COLLAB \\
    \midrule
    GCN & 20.37 $\pm$ 5.24 & 54.87 $\pm$ 11.20 & 51.37 $\pm$ 11.06 & 56.16 $\pm$ 11.07 & 50.62 $\pm$ 7.13 \\
    GAT & 15.90 $\pm$ 4.13 & 48.78 $\pm$ 18.46 & 51.20 $\pm$ 27.93 & 53.19 $\pm$ 20.61 & 51.08 $\pm$ 7.59 \\
    \midrule
    InfoGraph & 20.90 $\pm$ 3.32 & 54.12 $\pm$ 8.20 & 54.04 $\pm$ 9.45 & 57.57 $\pm$ 9.93 & 52.13 $\pm$ 8.71\\
    GraphCL & 28.11 $\pm$ 4.00 & 56.38 $\pm$ 7.24 & 55.40 $\pm$ 12.04 & 59.22 $\pm$ 7.42 & 52.81 $\pm$ 9.05\\
    \midrule
    CGPN & 24.75 $\pm$ 5.71 & 53.68 $\pm$ 9.15 & 52.16 $\pm$ 9.37 & 58.24 $\pm$ 8.47 & 50.05 $\pm$ 6.91 \\
    Meta-PN & 21.05 $\pm$ 4.58 & 54.17 $\pm$ 8.36 & 52.83 $\pm$ 10.26 & 56.37 $\pm$ 13.15 & 51.71 $\pm$ 7.12\\
    \midrule
    GPPT\tnote{1} & - & - & - & - & - \\
    GraphPrompt & 31.45 $\pm$ 4.32 & 64.42 $\pm$ 4.37 & 59.21 $\pm$ 6.82 & 61.63 $\pm$ 7.68 & 55.16 $\pm$ 6.24\\
    GPF & 32.65 $\pm$ 5.73 & 57.16 $\pm$ 5.96 & 61.62 $\pm$ 7.47 & 59.17 $\pm$ 6.18 & 53.91 $\pm$ 8.25 \\
    ProG & 29.18 $\pm$ 3.09 & 60.98 $\pm$ 7.49 & 61.96 $\pm$ 6.35 & 63.71 $\pm$ 5.25 & 54.93 $\pm$ 7.24 \\
    \midrule
    \name~ & \textbf{33.57 $\pm$ 4.72} & \textbf{64.95 $\pm$ 5.86} & \textbf{65.71 $\pm$ 5.34} & \textbf{68.58 $\pm$ 7.57} &
    \textbf{57.29 $\pm$ 6.17}\\
    \bottomrule
  \end{tabular}
  \begin{small}
  \begin{tablenotes}
      \item[1] GPPT\cite{sun2022gppt}~lacks a unified effort to address graph classification tasks.
    \end{tablenotes}
    \end{small}
  \end{threeparttable}
  }
\end{table}

\subsection{Model Analysis}
\label{sec:model_analysis}
We further analyse several aspects of our model.
The following experiments are conducted on the 3-shot node classification and 5-shot graph classification.

\begin{table}[t]
  \caption{Varying the ratio of added edges on few-shot node classification. $\downarrow$/$\uparrow$ means the decreasing/improvement compared with the accuracy of runner-up.}
  \setlength{\tabcolsep}{2.8pt}
  \label{tab:edge_ratio}
  \centering
  \resizebox{\textwidth}{!}{
  \begin{tabular}{c|c|cccccccc}
    \toprule
     & runner-up & 0\% & 0.1\% & 1\% & 5\% & 10\% & 50\% & 100\% \\
    \midrule
    Cora & 65.89$\pm$3.45 & 65.06$\pm$3.77 $\downarrow$ & 65.72$\pm$3.14 $\downarrow$ & 67.50$\pm$2.38 $\uparrow$ & 67.46$\pm$2.45 $\uparrow$ & 67.88$\pm$2.83 $\uparrow$ & 68.62$\pm$2.67 $\uparrow$ & 68.68$\pm$2.17 $\uparrow$\\
    PubMed & 70.07$\pm$6.07 & 70.84$\pm$5.31 $\uparrow$ & 70.32$\pm$3.95 $\uparrow$ & 71.62$\pm$4.34 $\uparrow$ & 70.94$\pm$4.40 $\uparrow$ & 71.42$\pm$4.70 $\uparrow$ & 72.24$\pm$4.33 $\uparrow$ & 73.23$\pm$4.20 $\uparrow$ \\
    Chameleon & 29.18$\pm$4.53 & 25.22$\pm$3.09 $\downarrow$ & 30.92$\pm$3.07 $\uparrow$ & 33.41$\pm$3.38 $\uparrow$ & 33.30$\pm$2.36 $\uparrow$ & 33.07$\pm$3.79 $\uparrow$ & 32.09$\pm$2.20 $\uparrow$ & 33.97$\pm$3.22 $\uparrow$ \\
    \bottomrule
  \end{tabular}}
\end{table}

\begin{figure}[t]
    \centering
    \includegraphics[width=0.68\textwidth]{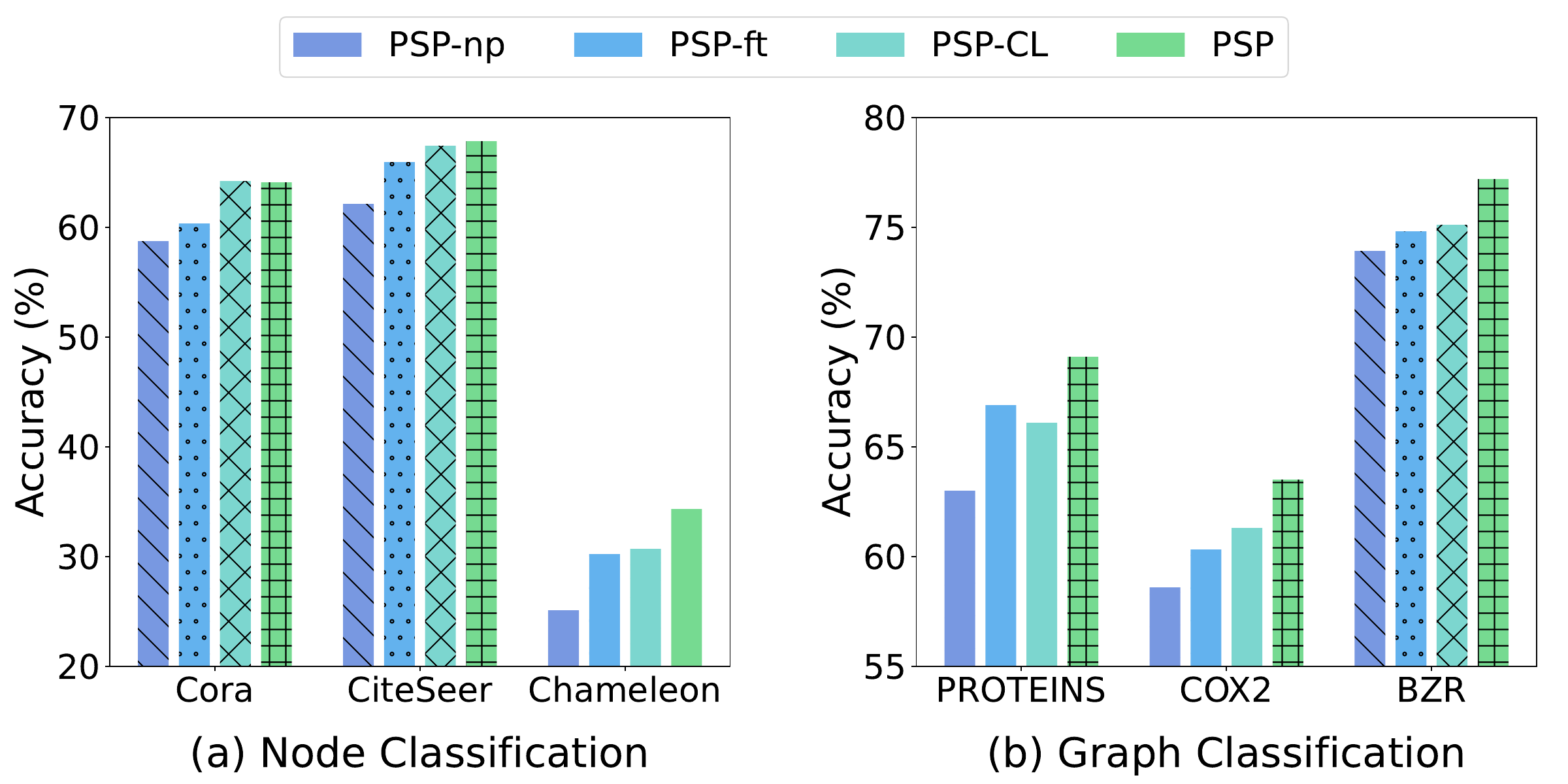}
    \caption{The ablation study on training paradigm.}
    \label{fig:ablation}
\end{figure}

\noindent \textbf{Ablation study on training paradigm.}
We conduct an ablation study that compares variants of \name~with different pre-training and prompt tuning strategies:
(1) We directly fine-tune the pre-trained models on the tasks, instead of prompt tuning.
We call this variant \textbf{\name-ft} (with \textbf{f}ine-\textbf{t}une).
(2) For downstream tasks,
we remove the prompt tuning process and only use the mean embedding vectors of the labeled data as the prototype vectors to perform classification.
We call this variant \textbf{\name-np} (\textbf{n}o \textbf{p}rompt).
(3) We replace our proposed dual-view contrastive learning in pre-training with GraphCL,
i.e.,
the pre-trained encoder is GNN,
while the downstream tasks still use our proposed graph structure prompt.
We call this variant \textbf{\name-CL} (with Graph\textbf{CL} as the pre-training model).
Figure \ref{fig:ablation} shows the results of this study, we observe that

\noindent (1) \name-np always performs the worst among all the variants,
showing the effectiveness of our proposed graph structure prompt.
\name-ft achieves better performance than \name-np,
which is because \name-ft is parameterized.

\noindent (2) \name~achieves comparable results to \name-CL in homogeneous graph and clearly outperforms \name-CL in heterophilous graph.
This is because our pre-training method uses MLP and GNN for contrastive learning,
where MLP is not affected by the heterophily of the graph.
In contrast, \name-CL is susceptible to structure heterophily.

\begin{figure}[t]
  \begin{subfigure}{0.25\linewidth}
    \centering
    \includegraphics[width=\linewidth]{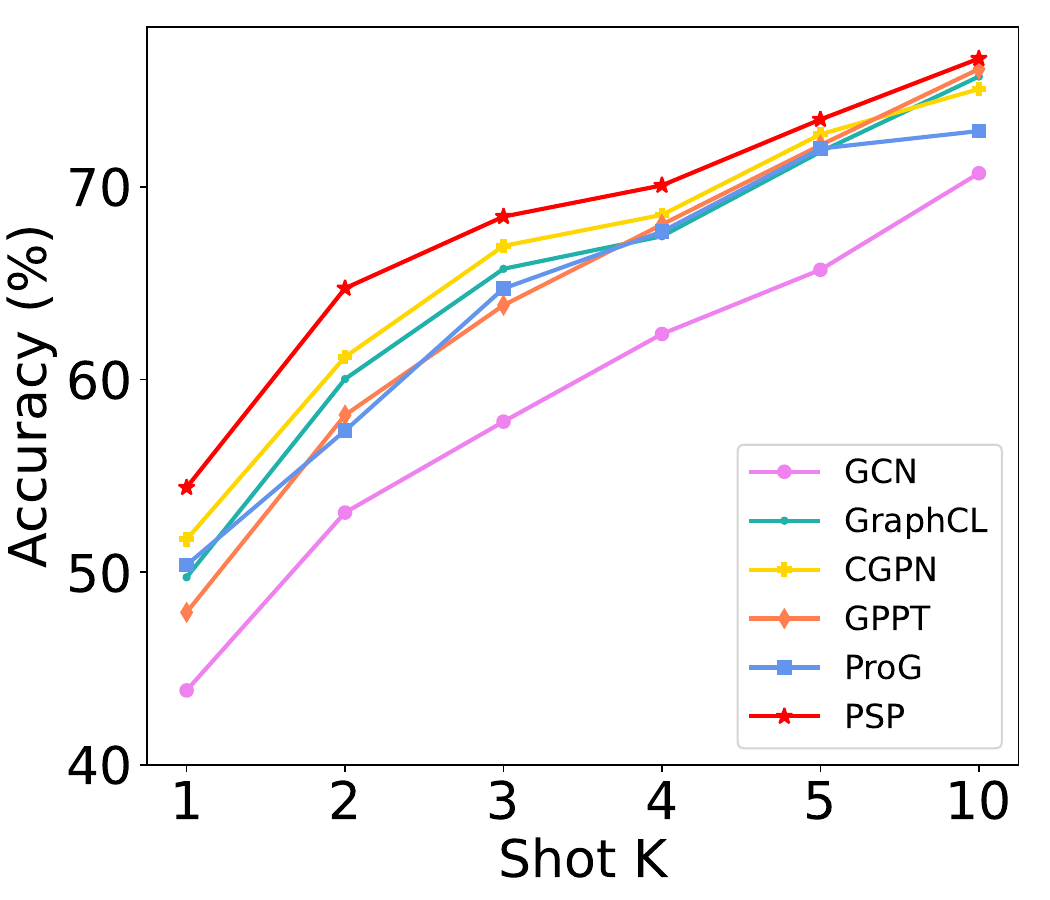}
    \caption{Cora}
    \label{fig:kshot_cora}
  \end{subfigure}
  \hspace{-2mm}
  \begin{subfigure}{0.25\linewidth}
    \centering
    \includegraphics[width=\linewidth]{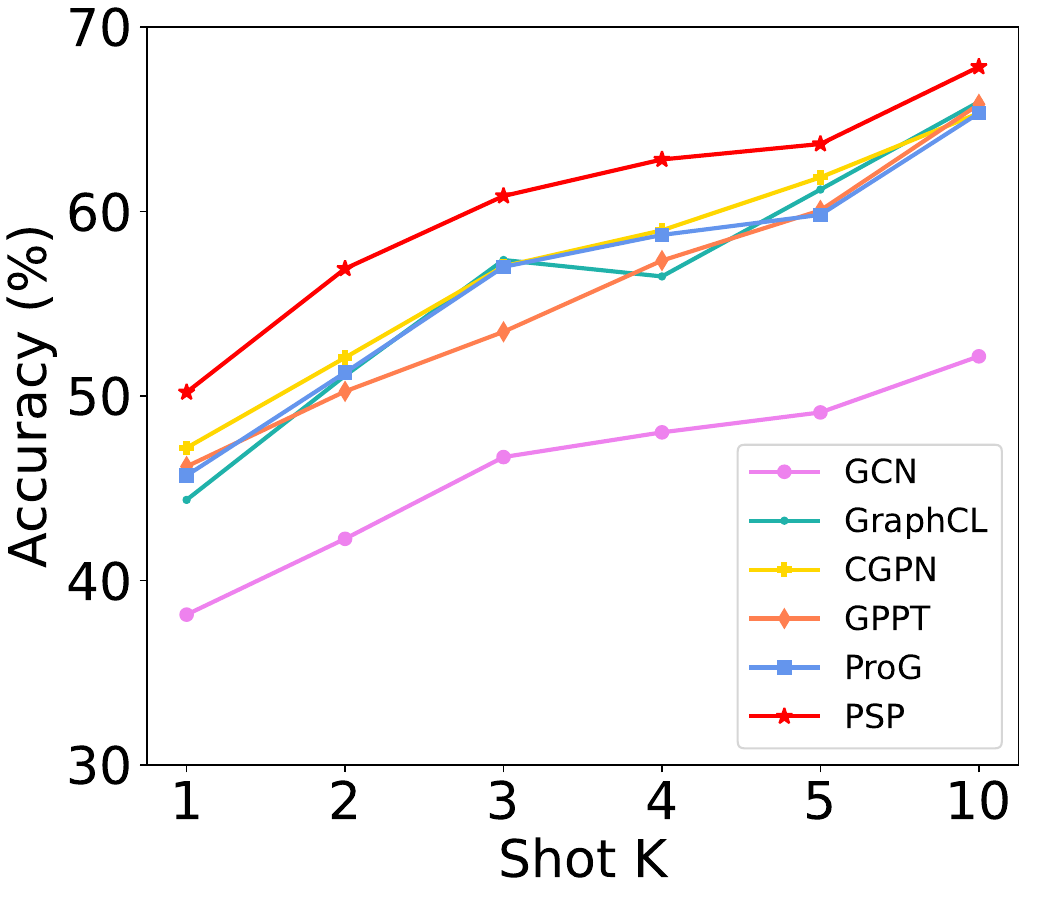}
    \caption{CiteSeer}
    \label{fig:kshot_citeseer}
  \end{subfigure}
    \hspace{-2mm}
    \begin{subfigure}{0.25\linewidth}
    \centering
    \includegraphics[width=\linewidth]{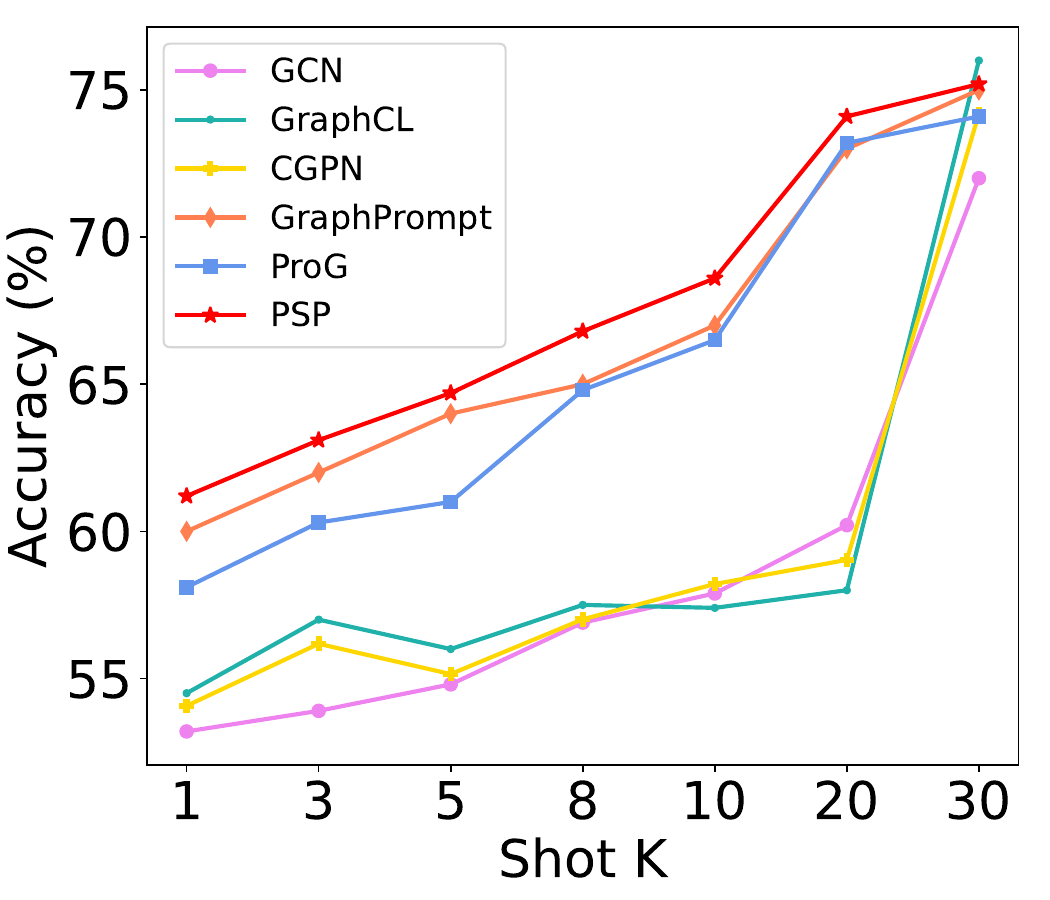}
    \caption{PROTEINS}
    \label{fig:kshot_proteins}
  \end{subfigure}
  \hspace{-2mm}
  \begin{subfigure}{0.25\linewidth}
    \centering
    \includegraphics[width=\linewidth]{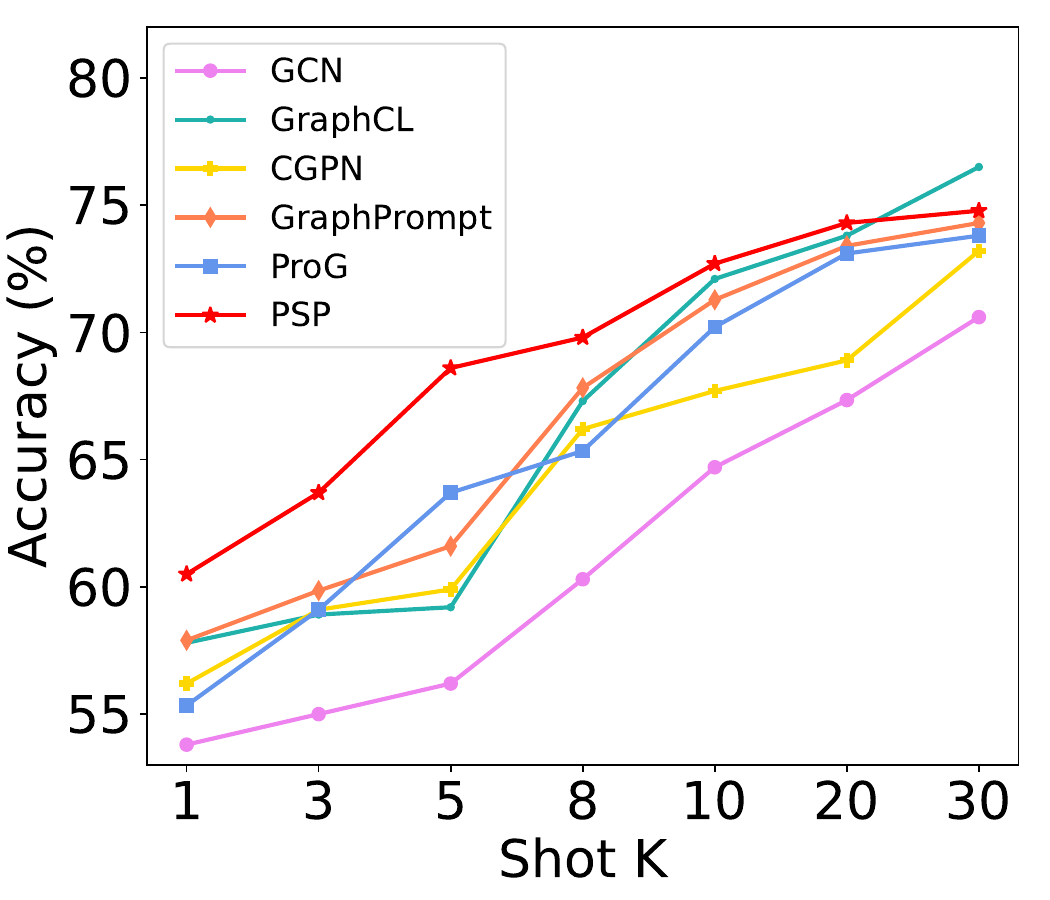}
    \caption{COX2}
    \label{fig:kshot_cox2}
  \end{subfigure}
  \caption{(a)(b): varying the number of shots for node classification. (c)(d): varying the number of shots for graph classification.}
  \label{fig:kshot_node}
\end{figure}

\noindent \textbf{Varying the number of the shots.}
For node classification,
we vary the number of shots between 1 and 10.
For graph classification,
we vary the number of shots between 1 and 30.
We compare \name~with several competitive baselines in Figure \ref{fig:kshot_node}.
In general, \name~consistently outperforms the baselines,
especially when the number of shots is few. 
We further notice that when the number of shots is relatively large,
\name~can be surpassed by graphCL on graph classification,
especially on COX2.
This could be contributed to more available training data on COX2,
where 30 shots per class implies that 12.85\% of the 467 graphs are used for training.
This is not our target few-shot scenario.

\noindent \textbf{Varying the ratio of added edges.}
Our prompt tuning method is parameterized by
the added edges between original nodes and prototype vectors,
where the weights of added edges are learnable.
We hereby study the impact of the ratio of newly added
edges (i.e., parameters) $r$.
To vary the number of edges, we first
randomly select $rN$ nodes outside the training set.
During prompt tuning, 
{we next combine the $rN$ nodes} with $N_t$ training nodes as the set of nodes that add connections with all the $C$ prototypes.
Following this setting, we conduct 3-shot node classification on 4 datasets. The results are shown in Table \ref{tab:edge_ratio}.
Surprisingly,
we observe that
for Cora and Chameleon,
\name~can surpass the runner-up when $r=1\%$ and $0.1\%$, respectively.
For PubMed,
\name~can outperform the runner-up even when $r=0$.
This shows that our prompt tuning model can achieve superior performance with only a small number of parameters $(N_t+rN)C$, where $r$ is a small number.
Therefore, \name~is promising in scaling to large graphs.

\begin{figure}[t]
    \centering
    \includegraphics[width=0.75\textwidth]{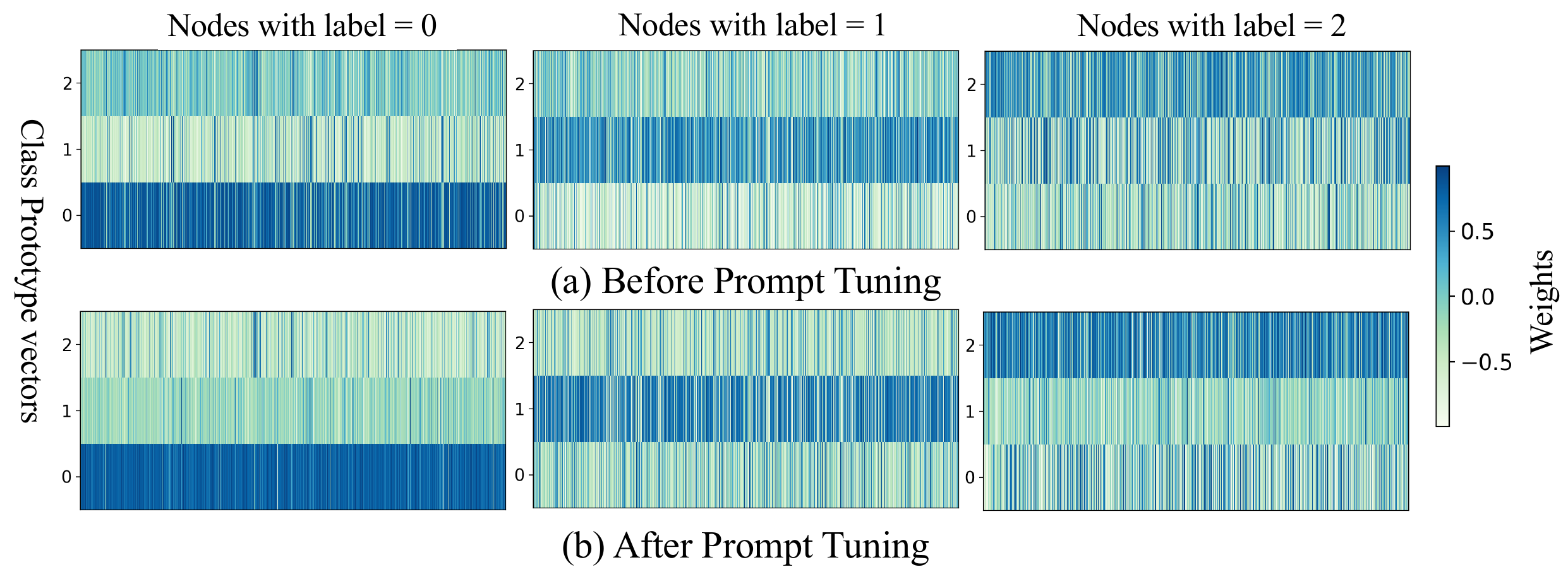}
    \caption{The weights of added edges between nodes and class prototype vectors before and after prompt tuning.}
    \label{fig:prompt}
\end{figure}

\noindent\textbf{Visualization of learned edge weights}. 
We visualize the weight matrix $\mathbf{W}$ of the added edges after prompt tuning. As shown in Figure \ref{fig:prompt},
for each node, its edge connected to the corresponding class prototype is {more likely to} have a larger weight.
Hence, the prototype vectors are very accurate by aggregating massive unlabeled data which contains rich pre-trained knowledge to reflect the semantics of the task labels.


\section{conclusion}

In this paper, we proposed \name,
a novel pre-training and structure prompt tuning framework for GNNs,
which unified the objectives of pre-training and prompt tuning for GNNs and integrated structural information in both pre-training and prompt tuning stages {to construct more accurate prototype vectors}.
For pre-training,
we proposed a dual-view contrastive learning to align the latent semantic spaces of node attributes and graph
structure.
For downstream prompt tuning,
we proposed to learn the structural connection between the prototype vectors and the graph,
and then leveraged the learned structural information to perform better in few-shot tasks.
Finally, we conducted extensive experiments and showed that \name~significantly outperforms various state-of-the-art baselines on both homophilous and heterophilous graphs, especially on few-shot scenarios.

\section{Acknowledgement}
This work is supported by National Natural Science Foundation of China No. 62202172 and Shanghai Science and Technology Committee General Program No. 22ZR1419900.


\end{document}